\documentclass[sigconf]{aamas} 

\usepackage{microtype}
\usepackage{graphicx} 
\usepackage{amsmath}
\usepackage{multirow}
\usepackage{booktabs} 
\usepackage{caption}

\usepackage{balance} 

\usepackage{natbib}
\setcitestyle{numbers,square}

\setcopyright{ifaamas}
\acmConference[AAMAS '23]{Proc.\@ of the 22nd International Conference
on Autonomous Agents and Multiagent Systems (AAMAS 2023)}{May 29 -- June 2, 2023}
{London, United Kingdom}{A.~Ricci, W.~Yeoh, N.~Agmon, B.~An (eds.)}
\copyrightyear{2023}
\acmYear{2023}
\acmDOI{}
\acmPrice{}
\acmISBN{}

\acmSubmissionID{828}

\title[]{Think Twice: A Human-like Two-stage Conversational Agent for Emotional Response Generation}

\author{Yushan Qian}
\affiliation{
  \institution{Tianjin University}
  \city{Tianjin}
  \country{China}}
\email{yushanqian@tju.edu.cn}

\author{Bo Wang}
\affiliation{
  \institution{Tianjin University}
  \city{Tianjin}
  \country{China}}
\email{bo_wang@tju.edu.cn}

\author{Shangzhao Ma}
\affiliation{
  \institution{Tianjin University}
  \city{Tianjin}
  \country{China}}
\email{shangzhaoma@tju.edu.cn}

\author{Wu Bin}
\affiliation{
  \institution{Quesoar Co. Ltd.}
  \city{Tianjin}
  \country{China}}
\email{wubin@quesoar.com}

\author{Shuo Zhang}
\affiliation{
  \institution{Quesoar Co. Ltd.}
  \city{Tianjin}
  \country{China}}
\email{s@quesoar.com}

\author{Dongming Zhao}
\affiliation{
  \institution{Artificial Intelligence Laboratory, China Mobile Communication Group Tianjin Co., Ltd.	}
  \city{Tianjin}
  \country{China}}
\email{waitman_840602@163.com}

\author{Kun Huang}
\affiliation{
  \institution{Artificial Intelligence Laboratory, China Mobile Communication Group Tianjin Co., Ltd.	}
  \city{Tianjin}
  \country{China}}
\email{hknzh@126.com}

\author{Yuexian Hou}
\affiliation{
  \institution{Tianjin University}
  \city{Tianjin}
  \country{China}}
\email{yxhou@tju.edu.cn}

\begin{abstract}
Towards human-like dialogue systems, current emotional dialogue approaches jointly model emotion and semantics with a unified neural network. 
This strategy tends to generate safe responses due to the mutual restriction between emotion and semantics, and requires the rare large-scale emotion-annotated dialogue corpus.
Inspired by the "think twice" behavior in human intelligent dialogue, we propose a two-stage conversational agent for the generation of emotional dialogue. Firstly, a dialogue model trained without the emotion-annotated dialogue corpus generates a prototype response that meets the contextual semantics. Secondly, the first-stage prototype is modified by a controllable emotion refiner with the empathy hypothesis. Experimental results on the DailyDialog and EmpatheticDialogues datasets demonstrate that the proposed conversational agent outperforms the compared models in the emotion generation and maintains the semantic performance in the automatic and human evaluations.
\end{abstract}


\newcommand{\BibTeX}{\rm B\kern-.05em{\sc i\kern-.025em b}\kern-.08em\TeX}

\begin{document}


\pagestyle{fancy}
\fancyhead{}


\maketitle 


\section{Introduction}

In the task of open-domain dialogue generation, emotional dialogue aims to generate responses involving the perception and expression of proper emotions. A large number of studies \citep{1,2,3} have demonstrated that emotional dialogue can significantly improve users’ satisfaction in a human-machine conversation. Moreover, building a dialogue system with human emotions is one of the ultimate goals of artificial intelligence.

Towards emotional dialogue systems, in addition to the early methods of manually compiling rules by professionals~\cite{4}, existing statistical approaches are mainly based on neural network models~\cite{5, 6, 7, 8, 9, 10, 11, 12, 13, 46}. With an end-to-end strategy, these neural network models jointly generate the semantics and emotions of the dialogue responses.

\begin{figure}
    \center
    \includegraphics[width=0.48\textwidth]{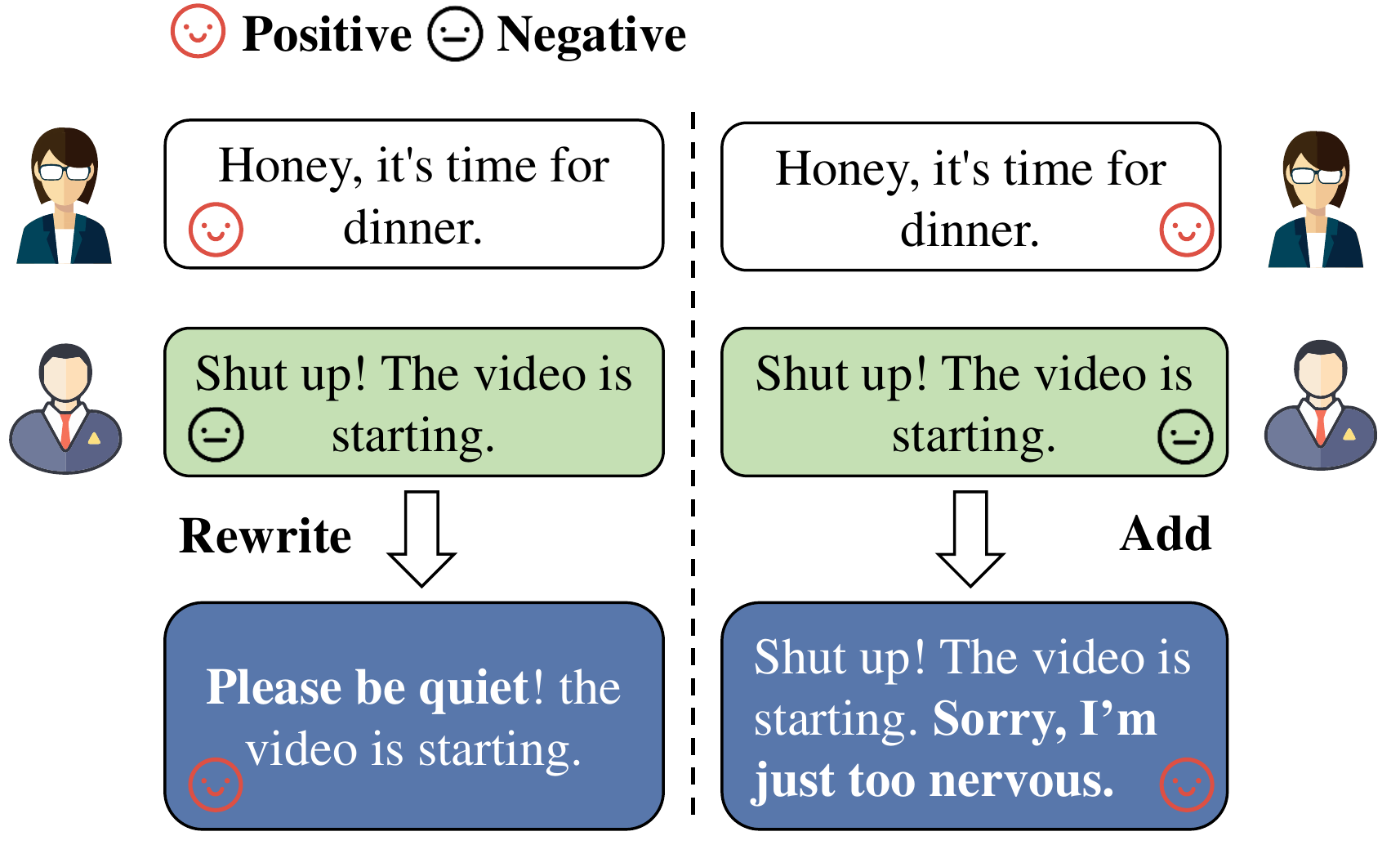}
    \caption{The real-life examples of emotion adjustment in the human dialogue. The "think twice" strategy can be observed in human intelligent behavior and effectively improves the quality of emotional responses by rewriting expressions or adding extra information. Bold tokens are the rewritten or added part.}
    \label{fig:1}
\end{figure}

However, current end-to-end emotional dialogue models still face several challenges. Firstly, in deep neural networks, the input emotion signals are often weakened through the complex learning process. Secondly, in the joint generation model, the design to enhance emotions often restricts the semantic performance of generated responses (e.g., safe responses). Thirdly, large-scale emotion-annotated dialogue corpora are rare for joint training of semantics and emotions with deep neural networks.

\begin{figure*}
    \centering
    \includegraphics[scale=0.5]{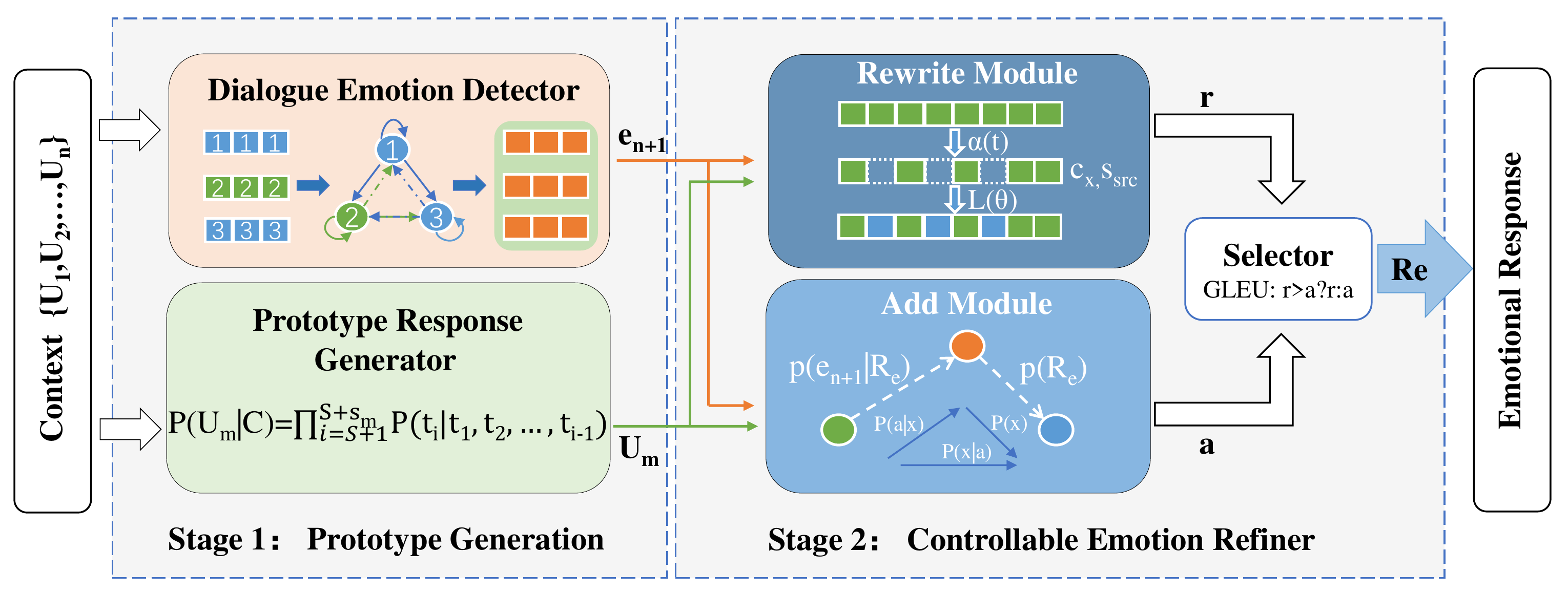}
    \caption{The overall architecture of the proposed two-stage conversational agent. The first stage includes Prototype Utterance Generator and Dialogue Emotion Detector, which generates prototype response $U_m$ and detects the contextual emotion state as the expected emotion $e_{n+1}$ of the final response, respectively. The second stage includes Rewrite Module, Add Module, and Selector. Rewrite Module and Add Module refine $U_m$ according to $e_{n+1}$, and the Selector selects the final response from the outputs of Rewrite Module and Add Module based on the GLEU score.}
    \label{fig:2}
\end{figure*}

In response to the above challenges, we propose to generate emotional responses with the idea of human intelligent dialogue behavior. When humans respond in a dialogue, the simultaneous processing of emotion and semantics can not ensure satisfying results~\cite{frijda1987emotion}. The intuitive emotion generated simultaneously with semantics is often arduous to ensure a response in the appropriate emotion. One source of the appropriate emotional response comes from an independent emotion selection after determining the semantics, i.e., thinking twice about appropriate emotions. In this independent emotion selection, a paramount strategy for humans to determine the appropriate emotion is the empathy strategy, which makes the emotion of the response consistent with that of the context~\cite{14, 11, 27}. As visualized in Figure~\ref{fig:1}, each response firstly has a proper semantic to respond to the context. Then, by recognizing the context’s emotional state, we can adjust the response emotionally and achieve a certain degree of empathy by responding to the partner’s emotion.

Therefore, we design a human-like two-stage conversational agent for emotional response generation. Firstly, a prototype response with proper semantics is generated with a pre-trained model fine-tuned on dialogue corpus without emotion annotation. Then, the contextual emotional state is recognized by a dialogue emotion detector. According to the empathy hypothesis~\cite{11}, the type of generated emotion is consistent with the contextual emotional state. Finally, the prototype is modified by a controllable emotion refiner to generate a final response that is both semantically relevant and emotionally appropriate. 

Specifically, towards effective refining for an emotional response in the second stage, we also refer to two human pragmatic strategies. First, humans express the same information in different ways with different vocabulary choices \citep{15}. Therefore, we involve expected emotional attributes in the response by replacing the original emotional words or phrases instead of constructing a new sentence from scratch, i.e., the "rewriting" strategy. Second, there are also some implicit emotions reflected through the whole sentence instead of specific words. Consequently, we also adjust the emotion by adding extra sentences to the response, i.e., the "adding" strategy.

In summary, our contributions are:
\begin{itemize}
\item	Inspired by human intelligent dialogue behavior, we propose a human-like two-stage conversational agent for emotional response generation. To the best of our knowledge, it is the first two-stage model specifically for emotional dialogue.

\item	The proposed method effectively alleviates two problems of existing emotional dialogue approaches, i.e., weakening the emotion effect during the complex learning process and restricting the semantic generation to meet the emotion demand.

\item	The proposed two-stage conversational agent reduces the demand for the sizeable emotion-annotated dialogue corpus. The training of the prototype response generator in the first stage only requires general dialogue corpora without emotion annotation, and the controllable emotion dialogue refiner is trained on non-dialogue and non-parallel emotion-annotated corpora.

\item	The proposed method can be generalized to other existing end-to-end emotion dialogue generation models as post-processing for emotionalization. Even if some sentences have poor emotional expressions, there is no need to retrain the whole model and build new sentences from scratch.
\end{itemize}

\section{Methods}
\subsection{Preliminaries}
Formally, in this paper, the dialogue context is alternate utterances of two speakers, defined as $C=\left \{ U_1, U_2,\dots, U_n \right \}$, where $n$ denotes the number of utterances in a dialogue. The set of context emotions is $E=\left \{e_1, e_2,\dots,e_n \right \}$, which corresponds to the dialogue context $C$.  Our goal is to generate the next utterance $U_{n+1}$, which is coherent to the context and contains the appropriate emotion.

As shown in Figure~\ref{fig:2}, our model consists of three parts: the Prototype Utterance Generator $G$, the Dialogue Emotion Detector $D$, and the Controllable Emotion Refiner $R$. The Controllable Emotion Refiner $R$ has two modules, named "Rewrite" and "Add". The Prototype Utterance Generator $G$ takes the dialogue context $C$ as input and generates a prototype response $U_m$. The Dialogue Emotion Detector $D$ takes the dialogue context $C$ as input and obtains the emotion state set $E$, which dynamically determines the response emotion $e_{n+1}$. The Controllable Emotion Refiner $R$ refines $U_m$ according to $e_{n+1}$ by rewriting $U_m$ or adding extra sentences into $U_m$ with Rewrite Module and Add Module, respectively, and generates the final response $R_e$, which is $U_{n+1}$.

\subsection{Prototype Utterance Generator}
We use DialoGPT~\cite{18} as the Prototype Utterance Generator to generate relevant, diverse, and contextually consistent responses. Large-scale pre-trained language models~\cite{16,17,18} have extensively promoted the research progress of the open-domain dialogue in recent years. The method of pre-training and fine-tuning can avoid training models from scratch, save computing resources, and achieve excellent results in downstream tasks.

DialoGPT has a 12-to-48 layer Transformer with layer normalization like GPT-2~\cite{16}, which is trained on the 147M large-scale dialogue dataset from Reddit. All utterances of the context are spliced into a long sentence with ``\textless\textbar endoftext \textbar\textgreater'' as input. The conditional distribution of the target prototype utterance $U_m$ is the product of a series of conditional probabilities:
\begin{eqnarray}
    P(U_m|C)=  {\textstyle \prod_{i=S+1}^{S+s_m}P(t_i|t_1,t_2,...,t_{i-1})},
\end{eqnarray}
where $ C=\left \{ U_1,U_2,...,U_n \right \}$, $n$ is the number of utterances, $U_i=\left \{t_1,t_2,...,t_{s_i}\right \}$, $t_i$ is each token in the utterance, $s_i$ is the length of each utterance, $S=s_1+s_2+\cdots+s_n$ represents the number of tokens in all utterances, $s_m$ is the length of $U_m$. 

We use the maximum mutual information scoring function (MMI) and the top-k sampling~\cite{45} to reduce the generation of meaningless responses. The specific implementation is based on tools provided by Hugging Face\footnote{https://huggingface.co/models}. We conduct fine-tuning on the training set for 3 epochs with the batch size of 8 and the learning rate of $0.00001$. Emotional labels in the training set will not be used. During the decoding process, we use the top-k (k=100) sampling and nucleus sampling (p=0.7)~\cite{44}.

\subsection{Dialogue Emotion Detector}
As an intuitive hypothesis of empathy, during emotional dialogues between two individuals, the listener usually tends to respond in a way that recognizes the speaker’s feelings~\cite{14,11} and achieves a certain degree of empathy by calling the respondent’s emotions. In the dialogue scene of this work, there are not many turns in the dialogue context (<4), so we adopt this empathy hypothesis to determine the emotion of the response according to the emotion state of the context.

To this end, the goal of the Dialogue Emotion Detector $D$ is to detect emotions in the dialogue context. According to the empathy hypothesis, $D$ determines the expected emotion of the Controllable Emotion Refiner by the recognized emotion distribution in the dialogue context.

The Dialogue Emotion Detector $D$ is developed based on DialogueGCN~\cite{20}, which regards each utterance in the dialogue as a node of the graph network. There are directed edges between utterances, and the sequence order of utterances determines the direction of each edge. These directed edges can model the emotional impact of what the speaker or the other people has said before. We use Glove embedding and CNN to extract features of utterances and get the embedding of each utterance $U_i$, which is the vector of each node. There are $b$ utterances before each utterance, and there are $a$ utterances after it. The node of each utterance has edges with $a + b + 1$ nodes (including itself). The weight $a_{ij}$ of each edge is decided by the relationship between nodes as follows:
\begin{eqnarray}
    \begin{split}
        \alpha _{ij} = {\rm softmax}(U_{i}^{T}·W_u[U_{i-b},...,U_{i+a}]), \\
        \text{for\quad} j = i-b,...,i+a.
    \end{split}
\end{eqnarray} 

Further details about DialogueGCN construction are available in~\cite{20}. Finally, the embedding from the sequence encoder $sq$ and the speaker-level encoder $sp$ are spliced together, and combined with the similarity-based attention mechanism to obtain the final embedding of the utterance node. Then we use a fully connected network to classify multiple emotion categories:
\begin{equation}
    \begin{aligned}
        \qquad & H =[h_1,h_2,\dots,h_n], \\
        \qquad & h_i = {\rm softmax}([sq_i,sp_i]^TWH)H^T,\\
         & e_i  =  {\rm argmax}({\rm softmax}({\rm FFN}(h_i))).
    \end{aligned}
\end{equation}
We use L2 regularization classification cross-entropy loss as the loss function and Adam~\cite{21} as the optimizer. We classify emotions in the emotion state set $S$ into two groups of negative emotions and positive emotions as in~\cite{13}. Following~\cite{14,11}, we assume that empathetic responses may mimic the user’s emotions to some extent. Therefore, the target emotion $e_{n+1}$ that we finally pass to the Controllable Emotion Refiner $R$ is defined as follows:
\begin{equation}
    \begin{aligned}
        e_{n+1} &= {\rm positive },\\
        \text{\quad if\quad} Num_{pos}(E) &> Num_{neg}(E),\\
        \text{otherwise\quad} e_{n+1} &= {\rm negative.}&\\
    \end{aligned}
\end{equation}
Where $E=\left \{e_1, e_2,\dots,e_n \right \}$ is the set of emotions in each dialogue. $Num_{pos}$ and $Num_{neg}$ represent the number of positive emotions and negative emotions in $E$, respectively.

\subsection{Controllable Emotion Refiner}
The Controllable Emotion Refiner $R$ takes the prototype response $U_m$ and the target emotion $e_{n+1}$ as input, and generates the final response $R_e$. The goal we need to learn is defined as:
\begin{equation}
P(R_e|U_m,e_{n+1})\ \&\ {\rm Stype}(R_e)=e_{n+1},
\end{equation}
where ``Stype'' represents the emotion type.

The Controllable Emotion Refiner $R$ consists of two modules, ``Rewrite'' and ``Add''. The Rewrite Module transforms the emotion attribute of the $U_m$ by replacing the original emotion symbols in the sentence with symbols that express the target emotion. The Add module adjusts the emotion type by adding extra sentences. We train the Controllable Emotion Refiner $R$ by parts.

\textbf{Rewrite Module.} The Rewrite Module consists of two parts: the first one is the deletion part, which determines whether each token in the input is an emotion attribute word, learns the emotional part and non-emotional part in the input, and deletes the emotional part. We adopt the attention mechanism of Transformer to extract the attention score as the weight of each token~\cite{25}: 
\begin{eqnarray}
\alpha(t)={\rm softmax}(QK^T), for\ t \in U_m,
\end{eqnarray}
where $Q$ and $K$ carry the original connotations of query and key vectors in the 
Transformer.

The second is the generating part, which generates sentences with target emotion attributes. The generating part adopts the Transformer structure, based on the Hugging Face~\cite{22}. The input of generating part is the prototype response and the target emotion. The output is a sentence that conforms to the target emotion. Without requiring the parallel corpus, the training goal of generating part is to minimize the following reconstruction loss:
\begin{equation}
    \begin{split}
    L(\theta)=\sum_{x,s_{src} \in D} {\rm log}p(x|c_x,s_{src};\theta ),
    \end{split}
\end{equation}
where $D$ is the training dataset. Given a sentence $x$, the Rewrite Module model learns to reconstruct $y$ = $x$ with $c_x$, $s_{src}$. $c_x$ is the non-emotional content of $x$, and $s_{src}$ is the original style of the sentence.

\textbf{Add Module.} The Add Module is developed based on the work of~\cite{23} to change the emotion polarity of the original sentence by adding extra sentences with the target emotion. Using Bayes' theorem, we can use the model $p(x)$ and the model $p(a|x)$ to express the model $p(x|a)$:
\begin{equation}
    \begin{split}
    p(x|a) \propto p(a|x)p(x).
    \end{split}
\end{equation}

In order to obtain the required $p(x|a)$ to generate the sentence based on attribute $a$, we already have a language model $p(x)$ that can generate fluent sentences. Furthermore, we build a classifier to determine whether the text $x$ generated by the language model has $a$ attribute, that is, $p(a|x)$, then $p(x|a)$ can be obtained.

The process of the Add Module has three steps:

1. First, a forward pass is performed through the language model to compute the likelihood of the desired attribute using an attribute model that predicts $p(a|x)$.

2. Second, a backward pass updates the internal latent representations of the language model, using gradients from the attribute model to increase the likelihood of the sentence having the desired attribute.

3. Third, re-sampling to generate a new word according to the obtained new output probability distribution.

To generate more diverse sentences that conform to the language model, two methods are adopted to ensure that the language model of the generated sentence is as close as possible to the original language model: Kullback–Leibler (KL) Divergence and Post-norm Geometric Mean Fusion. About the language model, we use GPT2. Regarding the specific attribute discriminator $p(a|x)$, we take the existing non-dialogue emotion-annotated corpus and pre-train a classifier.

\textbf{Selector.} A Selector is designed to determine whether the response is from the Rewrite Module or the Add Module is selected as the final output. The Selector uses GLEU~\cite{24} as a basis for judging the overall effect of responses, which compares with the prototype response. \citet{25} found that GLEU is more suitable for human score than BLEU score. The Selector selects the final response with a higher GLEU score. 

\begin{table}
\caption{Two groups of emotions in the  DailyDialog dataset according to positivity and negativity.}
\centering
\begin{tabular}{p{9em}|p{9em}}
\hline
{Positive} & {Negative} \\
\hline
{happiness, surprise, other} & {anger, disgust, fear, sadness} \\
\hline
\end{tabular}
\label{tab:emo-group1}
\end{table}

\begin{table}
\caption{Two groups of emotions in the EmpatheticDialogues dataset according to positivity and negativity.}
\centering
\begin{tabular}{p{8.5em}|p{12em}}
\hline
{Positive} & {Negative} \\
\hline
{confident, joyful, grateful, impressed, proud, excited, trusting, hopeful, faithful, prepared, content, surprised, caring} & {afraid, angry, annoyed, anticipating, anxious, apprehensive, ashamed, devastated, disappointed, disgusted, embarrassed, furious, guilty, jealous, lonely, nostalgic, sad, sentimental, terrified} \\
\hline
\end{tabular}
\label{tab:emo-group2}
\end{table}

\section{Experiments}
In this section, we introduce the datasets, baselines and evaluation metrics. The proposed conversational agent is experimentally compared with baselines and the experimental results are discussed.

\subsection{Datasets}
We used the DailyDialog~\cite{26} and EmpatheticDialogues~\cite{27} datasets for the experiment. DailyDialog is a multi-round dialogue dataset for daily chat scenes. There are a total of 12,218 dialogues and 103,607 utterances. The topic and emotion in each utterance are labeled. There are seven types of emotion: anger, disgust, fear, happiness, sadness, surprise, and others. Refer to \citet{13}, we divided the 7 emotion types into two groups containing 3 positive and 4 negative emotions, respectively, as listed in Table~\ref{tab:emo-group1}. EmpatheticDialogues is a widely-used benchmark dataset for empathetic response generation, which is a large-scale multi-turn dataset containing 25k empathetic dialogues between crowdsourcing workers. EmpatheticDialogues also provides an emotion label for each dialogue from 32 available emotions. Following \citet{13}, we divided the 32 emotion types into two groups containing 13 positive and 19 negative emotions, respectively, as listed in Table ~\ref{tab:emo-group2}. We focus on positive and negative emotions because the consistency of polarity level emotion is more popular in emotion study and robust in the application. Since our method and baselines are under the same assumption and processed in the same way during evaluation, the results are competitive and convincing.

Considering the limited running space and to unify the number of rounds in each dialogue, we segment the original dialogues into sub dialogues having 4 rounds. Finally, for the DailyDialog dataset, the dialogue numbers of the training / validation / test set are 54,299 / 5,109 / 4,782, respectively. For the EmpatheticDialogues dataset, the dialogue numbers of the training / validation / test set are 18,383 / 2,810 / 3,320, respectively. 

\begin{table*}
\caption{Automatic and Human evaluations in the DailyDialog dataset (The significant improvement with p-value < 0.05 (t-test). Fleiss’s kappa for Human evaluation is 0.526, indicating "moderate agreement").}
\centering
\begin{tabular}{l|c|c|c|c|c|c|c|c}
\hline
    \multirow{2}{*}{Model} & 
    \multicolumn{4}{c|}{Automatic Evaluation} &
    \multicolumn{4}{c}{Human Evaluation} \\
    \cline{2-9}
     & {BLEU-4}
     & {Dist-1} 
     & {Dist-2} 
     & {Acc(\%)} 
     & {Con} & {Emo} & {Int} & {Flu}\\ \hline
    Transformer &  0.44 & 1.10 & 5.84 & 54.04 & 3.54  & 0.69 & 0.49 & 4.60\\
    Multi-TRS & 0.58 & 1.17 & 6.35 & 54.29 & 3.49  & 0.75 & 0.52 & 4.35 \\ 
    Mojitalk & 0.58	& 4.99 & 22.40	& 53.99	& 3.14 & 0.79 & 0.61 & 4.21\\
    MoEL	& 0.46 & 1.51	& 8.39 & 55.44 & 3.35 & 0.77 & 0.73 & \textbf{4.67} \\
    MIME	& 0.54 & 0.48	& 1.79 & 53.43 & 3.42 & 0.78 & 0.68 & 4.43\\
    EmpDG	& 0.57 & 1.12	& 5.73 & 56.02 & 3.40  & 0.79 & 0.77 & 4.47\\
    \hline
    Our & \textbf{0.67} & \textbf{9.71} & \textbf{42.77} & \textbf{57.36} & \textbf{3.82}  & \textbf{0.84} & \textbf{0.79} & 4.45  \\
   \hline
\end{tabular}

\label{tab:eval1}
\end{table*}

\begin{table*}
\caption{Automatic and Human evaluations in the EmpatheticDialogues dataset (The significant improvement with p-value < 0.05 (t-test). Fleiss’s kappa for Human evaluation is 0.462, indicating "moderate agreement").}
\centering
\begin{tabular}{l|c|c|c|c|c|c|c|c}
\hline
    \multirow{2}{*}{Model} & 
    \multicolumn{4}{c|}{Automatic Evaluation} &
    \multicolumn{4}{c}{Human Evaluation} \\
    \cline{2-9}
     & {BLEU-4}
     & {Dist-1} 
     & {Dist-2} 
     & {Acc(\%)} 
     & {Con} & {Emo} & {Int} & {Flu}\\ \hline
    Transformer &  0.35 & 0.64 & 2.40 & 62.29 & 3.24 & 0.62 & 0.84 & \textbf{4.86} \\
    Multi-TRS & 0.35 & 0.73 & 2.77 & 61.36	& 3.30  &  0.76 & 0.98 & 4.80\\ 
    Mojitalk & 0.23	& \textbf{6.99} & \textbf{33.52} & 61.81 & 2.74 & 0.76 & 1.03 & 4.54\\
    MoEL	& 0.34 & 1.15	& 7.28 & 62.47 & 3.44 & 0.77 & 1.04 & 4.82 \\
    MIME	& 0.37 & 0.89	& 3.93 & 61.87 & 3.32 & 0.80 & \textbf{1.12} & \textbf{4.86}  \\
    EmpDG	& \textbf{0.39} & 0.75	& 2.50 & 62.62 & 3.33 & 0.81 & 1.05 & 4.78 \\
    \hline
    Our & \textbf{0.39} & 5.24 & 22.37 & \textbf{65.42} & \textbf{3.65} & \textbf{0.82} & 1.11 &  4.72 \\
   \hline
\end{tabular}

\label{tab:eval2}
\end{table*}

\subsection{Compared Models}
To the best of our knowledge, this is an early work in the two-stage generation of emotional dialogue. In view of the empathy hypothesis, we compare our approach with a range of models used in related tasks, including general dialogue, emotional dialogue, and empathetic dialogue. 

\textbf{Transformer}~\cite{19}: The standard Transformer model that is trained to optimize NLL loss. 

\textbf{Multi-TRS}~\cite{27}: A multi-task Transformer model jointly trained by predicting the emotion and generating the response.

\textbf{Mojitalk}~\cite{8}: An encoder-decoder based CVAE model incorporated with emotion embedding. 

\textbf{MoEL}~\cite{11}: A Transformer-based model employs emotion-specific decoders whose outputs are aggregated and fed to a final decoder to generate the empathetic response.

\textbf{MIME}~\cite{13}: A Transformer-based model that leverages emotion groups and emotion mimicry, which effectively blends emotions in positive and negative emotion groups and generates the empathetic response.

\textbf{EmpDG}~\cite{46}: An interactive adversarial model consists of a generator and a discriminator. The discriminator requires user feedback. Besides, the model exploits both the coarse-grained dialogue-level and fine-grained token-level emotions. Referring to \citet{47}, we only apply the empathetic generator to ensure consistent input and output in the test set for a fair comparison with other baselines.

\subsection{Implementation Details}

We use the official codes of all baselines, especially, EmpDG only applies the empathetic generator (Multi-TRS\footnote{https://github.com/facebookresearch/EmpatheticDialogues}, Mojitalk\footnote{https://github.com/Claude-Zhou/MojiTalk}, MoEL\footnote{https://github.com/HLTCHKUST/MoEL}, MIME\footnote{https://github.com/declare-lab/MIME}, EmpDG\footnote{https://github.com/qtli/EmpDG}). We implement all the models using PyTorch except Mojitalk. All the baselines were trained on a V100 GPU with the batch size of 16 and the early stopping strategy. About the Adam optimizer, we set $\beta1$ = 0.9 and $\beta2$ = 0.98. For the emotion detection in the automatic evaluation, emotion pre-training model in Senta is ``$ernie\_2.0\_skep\_large\_en$''.

\subsection{Evaluation Metrics}
\subsubsection{Automatic Evaluation} 

We apply the following evaluation metrics in the automatic evaluation:

\textbf{BLEU}: Word-overlap scores with human responses~\cite{48}. We use BLEU-4, which is calculated with an NLG evaluation toolkit \footnote{https://github.com/Maluuba/nlg-eval}.

\textbf{Diversity}: Dist-n measures the proportion of unique n-grams in the generated responses~\cite{29}. It is commonly used to evaluate whether the dialogue model can generate a diverse response as humans do. Low diversity often means the model tends to generate similar safe responses to different contexts. We refer to the work of~\cite{28} to calculate the Dist-1 and Dist-2 metrics.

\textbf{Emotion Accuracy (Acc)}: The emotion accuracy is defined as the proportion of consistent emotion polarities between generated responses and the ground truth. We use the Sentiment Knowledge Enhanced Pre-training for Sentiment Analysis (SKEP) model~\cite{tian-etal-2020-skep} proposed by Baidu as the emotion detector during the evaluation\footnote{https://github.com/baidu/Senta}. SKEP is a state-of-the-art emotion detector in 14 typical Chinese and English sentiment analysis tasks. We use it to automatically detect the emotion polarity of the responses generated by our proposed conversational agent and baselines. 

\begin{table*}
\caption{ Ablation Analysis.}
\centering
\begin{tabular}{l|c|c|c|c|c|c|c|c}
\hline
    \multirow{2}{*}{Model} & 
    \multicolumn{4}{c|}{DailyDialog} &
    \multicolumn{4}{c}{EmpatheticDialogues} \\
    \cline{2-9}
     & {BLEU-4}
     & {Dist-1} 
     & {Dist-2} 
     & {Acc(\%)} 
     & {BLEU-4}
     & {Dist-1} 
     & {Dist-2} 
     & {Acc(\%)} \\ \hline
       Our & \textbf{0.67} & \textbf{9.71} & \textbf{42.77} & \textbf{57.36} & \textbf{0.39} & \textbf{5.24} & \textbf{22.37} & 65.42 \\
    \hline
    w/o Add	& 0.59 & 7.48	& 33.79 & 56.04 & 0.29 & 4.07 & 19.63 & 65.33 \\
    w/o Rewrite	& 0.49 & 7.93 & 35.74 & 55.88 & 0.35 & 4.95 & 20.27 & \textbf{65.78} \\
    w/o DED & 0.63 & 8.77 & 39.27 & 56.84 & — & — & — & — \\
   \hline
\end{tabular}
\label{tab:eval3}
\end{table*}

\subsubsection{Human Evaluation}
We randomly sampled 100 dialogues and generated responses with our proposed conversational agent and baselines. We employed three human annotators to evaluate each response based on four aspects:

\textbf{Content(Con)}: Whether the response is appropriate for the context in the current dialogue. It is rated on a Likert scale (1: not at all, 3: somewhat, 5: very much).

\textbf{Emotion(Emo)}: Whether the response is appropriate for the context in the emotion polarity. 1 indicates the response is appropriate, and 0 indicates the response is inappropriate.

\textbf{Emotion-intensity(Int)}: What the emotion intensity of the response is. 0 represents no emotion, 1 represents slight intensity, and 2 represents strong intensity.

\textbf{Fluency(Flu)}: Whether the response is readable and understandable. It is rated on a Likert scale (1: not at all, 3: somewhat, 5: very much).


\subsection{Main Results}
Both automatic and human evaluation results are shown in Table~\ref{tab:eval1} and Table~\ref{tab:eval2} on the DailyDialog and EmpatheticDialogues datasets, respectively.

For the performance of emotion generation, it can be observed that the proposed conversational agent outperforms baselines in \textbf{Acc} for the automatic evaluation and \textbf{Emo} for the human evaluation in two datasets, indicating the outstanding performance of our conversational agent in emotion generation. In addition, our conversational agent also achieves the best and second-best results in \textbf{Int} on the DailyDialog and EmpatheticDialogues datasets, respectively. which clearly verifies that the emotional effect of our model is more significant and sufficient compared with the SOTA end-to-end systems. This is because in the process of the two stages, the emotional effect of the response is separately determined and refined in the second stage without being influenced by the semantic generation in the first stage. 

For the performance of semantic generation, the proposed conversational agent reaches the highest level in \textbf{BLEU-4} and \textbf{Con}. In terms of \textbf{Dist-1} and \textbf{Dist-2}, our conversational agent also scores moderately. These results confirm that the proposed conversational agent significantly improves emotional expression while maintaining appropriate semantics. We also note that our proposed conversational scores are ordinary in \textbf{Flu}, which is possibly due to the "adding" strategy increasing the sentence length and affecting the reading difficulty. This is a small limitation given that all models score above 4, which we will explore in the future.

For the performance of compared baselines,  MoEL has a low \textbf{BLEU-4s} score and a high emotion accuracy score, which shows that this existing model loses semantic information when pursuing emotion features. Furthermore, all baselines except Mojitalk have low scores on diversity metrics, which indicates that there are a large number of safe responses in the generated responses. MoEL, Mojitalk, MIME, and EmpDG also have low scores in \textbf{Con} in the DailyDialog dataset, which is lower than Transformer. This may be primarily due to the mutual restriction of semantics and emotions, which reduces the output space. 

Since there are 8 metrics (Automatic and Human Evaluation) evaluated, which are less than 30, we choose t-test. Specifically, we use "ttest\_ind" in dcdd"scipy.stats" package to calculate metric values of our method with the metric values of baselines. All p-values are less than 0.05, accept the assumption, which means it is statistically significant.

\begin{figure}[h]
\flushleft 
    \includegraphics[scale=0.268]{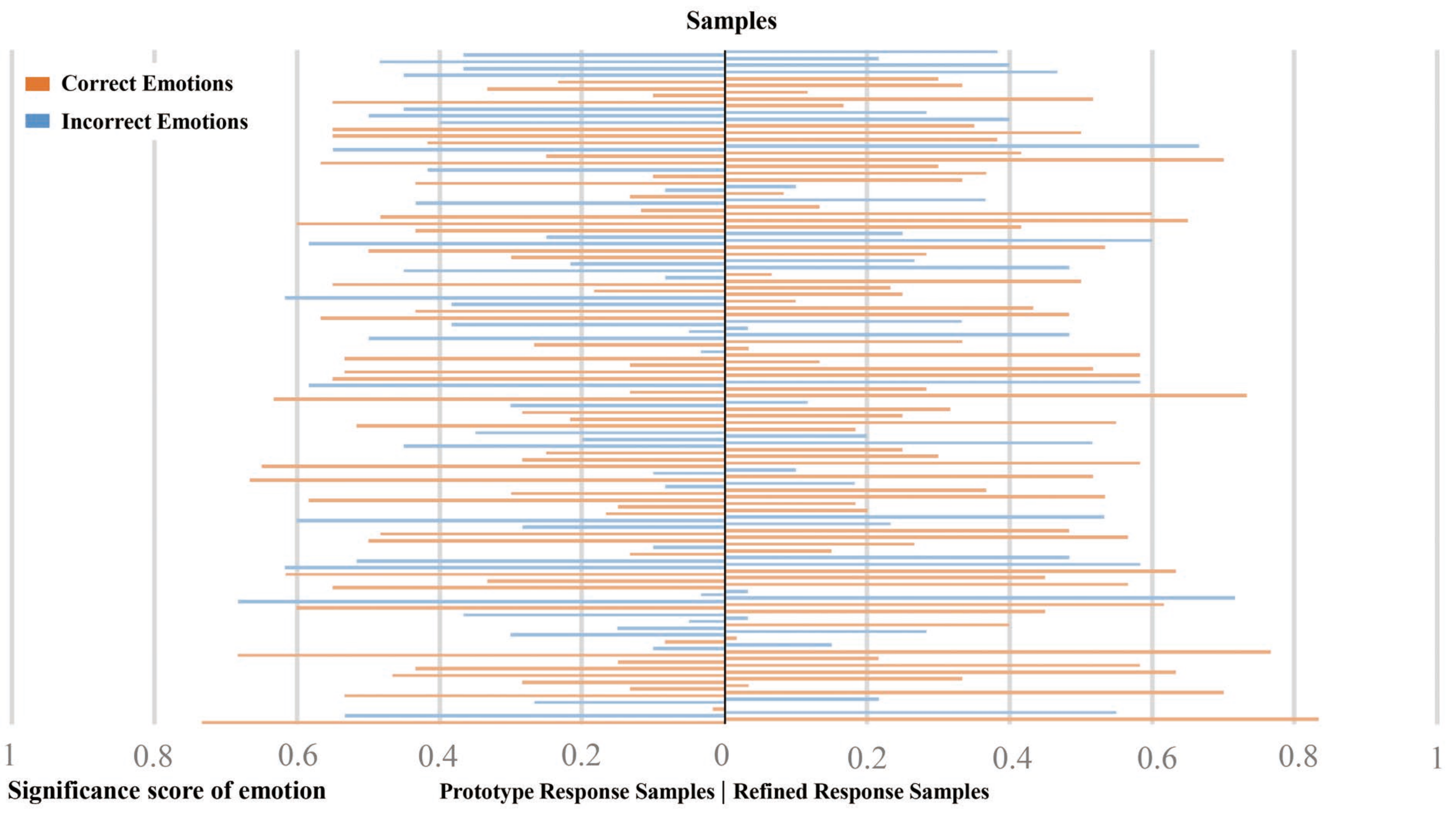}
    \caption{Compare the prototype response and refined response with respect to the correctness and significance of the emotion. The left and right columns indicate prototype and refined responses, respectively. The red and blue columns indicate correct (i.e., coherent to the contextual emotion)  and incorrect emotions, respectively. The length of the columns indicates the significance of the emotion.}
    \label{fig:3}
\end{figure}

In Figure~ \ref{fig:3}, for 100 human evaluation samples, we compared the correctness and significance of the emotion of the prototype response and the refined response generated by our proposed conversational agent. The red and blue columns indicate correct (i.e., coherent to the contextual emotion) and incorrect emotions, respectively. The length of the columns indicates the significance of the emotion. We can note that the number of red columns in the refined responses is more, and the length is longer, which illustrates that the refined responses improve the correctness and significance of the emotion in the prototype response.

\begin{table*}[] 
\caption{Examples of generated responses by our proposed conversational agent and baselines. The corresponding contexts did not appear in the training set. Bold tokens are the refined part. }
\label{tab:rst_table}
\resizebox{\textwidth}{!}{%
\begin{tabular}{*4{l}}
    \toprule  
	Context & Ground Truth Response & Models & Generated Response \\
	\midrule
	\multicolumn{4}{c}{DailyDialog}\\
	\hline
	\midrule
	\multirow{8}{*}{\begin{tabular}[c]{@{}l@{}}No, honey, why didn't you go to cram school\\ last night?\\ 
	Dad, I don't want to talk about it now. \\ 
	Honey, if you don't want to go to cram school, \\you should tell me the reason why.\end{tabular}} 
	& 
	\multirow{8}{*}{\begin{tabular}[c]{@{}l@{}}I'm sorry, dad. But I would\\ rather stay at school than go\\ to cram school.\end{tabular}} 
	& \multicolumn{1}{l}{Transformer} & I do not think I can do that . \\
	\cline{3-4}
	&  & Multi-TRS    & Yeah, I know, but I am sure I will.  \\
	\cline{3-4}
&  & Mojitalk    & Well, be able to be a good point.   \\
\cline{3-4}
& & Moel & I do not know. I want to go to school. \\
\cline{3-4}
& & MIME & That is true. But I do not know. \\
\cline{3-4}
& & EmpDG & I am sorry. I can not wait to get out of the internet. \\
\cline{3-4}
&& \textbf{Ours} & \textbf{Sorry} dad. I can't do that \textbf{again}. I'm so busy.\\
\hline
\bottomrule

	\multirow{7}{*}{\begin{tabular}[c]{@{}l@{}}
	\\I could have Suzy stay at my place. \\ 
	I don't want to do that to you, Karen.\\
	Why not? \end{tabular}} 
	& 
	\multirow{7}{*}{\begin{tabular}[c]{@{}l@{}}Your own kids would get \\ Suzy's cold. I think it would \\ be better to have a good\\ babysitter here. \end{tabular}} 
	& \multicolumn{1}{l}{Transformer} & Yes, I think I would love to. \\
	\cline{3-4}
& & Multi-TRS & I am sorry. I must have to give her a girl.\\
\cline{3-4}
&  & Mojitalk &  Well, I have a cold.  \\
\cline{3-4}
& & Moel &  I do not know. I would like to. You? \\
\cline{3-4}
& & MIME & I am not sure. I am not sure. \\
\cline{3-4}
& & EmpDG & I am sorry, but I have no idea. \\
\cline{3-4}
& & \textbf{Ours} & \begin{tabular}[c]{@{}l@{}} I'm sorry. I just can't. \textbf{I've got nothing to lose.}\\ \textbf{I just want to make the best of what I have, and the best I can.}\end{tabular} \\
\midrule

\multicolumn{4}{c}{EmpatheticDialogues}\\
	\hline
	\midrule
	\multirow{7}{*}{\begin{tabular}[c]{@{}l@{}}Why does McDonalds always get my order wrong!\\ I said no lettuce!\\
	Report to their management. \\ 
	Sometimes I think the management don't even care!\end{tabular}} 
	& 
	\multirow{7}{*}{\begin{tabular}[c]{@{}l@{}}Just give it a trial.\end{tabular}} 
	& \multicolumn{1}{l}{Transformer} & I am sorry to hear that. I hope they win. \\
	\cline{3-4}
	& & Multi-TRS & I hope you get a better next time. \\
	\cline{3-4}
&  & Mojitalk    & That's a great idea! I'm sure you'll get his family, he has done issues as well.   \\
\cline{3-4}
& & Moel &  I hope everything works out! \\
\cline{3-4}
& & MIME &  I am sorry to hear that. I hope you find a new job.\\
\cline{3-4}
& & EmpDG &  I do not know what I do. \\
\cline{3-4}
& & \textbf{Ours} & It is a problem. I am not sure how to solve it. \textbf{I feel very sorry.} \\
\hline
\bottomrule

	\multirow{8}{*}{\begin{tabular}[c]{@{}l@{}}
	I visited an orphan once and it was such a surreal 
	\\moment. It made me realise how thankful I should\\ be for
	a family.\\
That must have been and intense and emotional \\ moment for
you.\\
Yes it definitely was. I went with intentions to \\ teach the kid a few things but I ended up learning \\ lots more from him. \end{tabular}} 
	& 
	\multirow{8}{*}{\begin{tabular}[c]{@{}l@{}}Uau, what a surreal moment. \\I always believe kids have a\\ lot to teach us. \end{tabular}} 
	& \multicolumn{1}{l}{Transformer} & I am sure he'll be a good person.   \\
	\cline{3-4}
& & Multi-TRS &  I am glad you are able to find a new one. \\
\cline{3-4}
&  & Mojitalk  & Me too. The original kids is really much now.  \\
\cline{3-4}
& & Moel &  I am glad you are on the right?  \\
\cline{3-4}
& & MIME & That is great. I am glad you are happy for you. \\
\cline{3-4}
& & EmpDG & I am sure you will do it. \\
\cline{3-4}
& & \textbf{Ours} & \begin{tabular}[c]{@{}l@{}} I'm glad you had such \textbf{a great} experience. \end{tabular} \\
\cline{3-4}
& & & \\
\hline
\bottomrule
\end{tabular}
}
\end{table*}

\subsection{Human A/B Test}
We conducted the human A/B test, which is shown in Table~\ref{tab:ab}. We randomly sampled 100 response pairs and asked 3 annotators to choose the preferred response based on the dialogue context. A tie is allowed if both are good or bad. The inter-annotator agreement is measured by Fleiss's kappa. One example needs to be judged six times. Pay about 43.12\$ for every 100 examples. We can observe that responses generated by the proposed conversational agent are preferred by annotators over those generated by other models, which indicates that responses with appropriate emotions and diversity are more attractive to users. 

\begin{table*}
\caption{Result of human A/B test. Fleiss’ kappa result for DailyDialog and EmpatheticDialogues is 0.612 and 0.496, indicating "substantial agreement" and "moderate agreement", respectively.}
\centering 
  \begin{tabular}{l|ccc|ccc}
    \toprule
    \multirow{2}{*}{Models} & 
    \multicolumn{3}{c|}{DailyDialog} &
    \multicolumn{3}{c}{EmpatheticDialogues} \\
   & Win & Lose & Tie & Win & Lose & Tie \\
    \midrule
    Our vs Transformer & 45.0\% & 34.7\% & 20.3\% & 56.3\% & 27.3\% &16.3\% \\
    Our vs Multi-TRS & 46.0\% & 31.0\% & 23.0\% & 52.3\% & 28.0\% & 19.7\%\\
    Our vs Mojitalk & 56.3\% & 26.0\% & 17.7\% & 53.3\% & 26.0\% & 20.7\%\\
    Our vs MoEL & 44.7\% & 31.0\% & 24.3\% &46.7\% & 19.3\% & 14.0\%\\
    Our vs MIME & 49.0\% & 31.7\% & 19.3\% & 51.0\% & 25.7\% & 23.3\%\\
    Our vs EmpDG & 42.0\% & 32.0\% & 26.0\% & 48.0\% & 31.7\% & 26.3\%  \\
  \bottomrule
\end{tabular}
\label{tab:ab}
\end{table*}

\subsection{Ablation Analysis}
In order to verify the effectiveness of our proposed conversational agent, we also conducted ablation studies. \textbf{1) w/o Add}: The Add module is removed in the Controllable Emotion Refiner. We only consider using "rewriting" to refine the prototype response; \textbf{2) w/o Rewrite}: The Rewrite module is removed in the Controllable Emotion Refiner, and we consider using "adding" to refine the prototype response; \textbf{3) w/o DED}: The Dialogue Emotion Detector is removed and replaced by emotion recognition of a single utterance. This is only conducted in the DailyDialog dataset because the emotion annotation of the EmpatheticDialogues dataset is dialogue-level. 

As shown in Table~\ref{tab:eval3}, we can observe that removing the Add Module or the Rewrite Module both causes a drop in most metrics. This suggests that combining the "rewriting" and "adding" strategy is beneficial to generating appropriate responses in line with the human language characteristics of both explicit and implicit expression. However, the Selector of the second stage can be improved, such as selecting which module to use in advance based on the prototype response and emotion. How to refine the prototype response by the “Rewrite” or “Add” module more effectively and reasonably is a problem worth exploring. Moreover, the dialogue emotion detector also plays an important role in emotional response generation, which is superior to concatenating the context into a long sentence or identifying a single utterance.

\subsection{Case Study}
We sampled some generated responses from all models in Table~\ref{tab:rst_table}. We can observe that responses generated by other baselines have emotional expressions, but the semantics are less appropriate in general. Although responses generated by Transformer are fluent, they often do not conform to the context. In contrast, the response of the proposed conversational agent not only inherits the contextual semantics but also involves rich and appropriate emotions at the same time. For example, the proposed conversational agent coherently transforms "It is a problem. I am not sure how to solve it." to "It is a problem. I am not sure how to solve it. I feel very sorry." 


\section{Related Work}
\textbf{Dialogue Emotion Recognition.} Different from the emotion recognition of independent sentences, emotions in dialogue should be recognized by the context. Used context typically includes history utterances~\cite{30}, history emotions~\cite{31} and mutual influence of speakers~\cite{32}. To model the context, utterances and speakers can be independently~\cite{32} or interactively~\cite{33} modeled by GRU. \cite{20} uses GCN to solve the problem of context propagation in existing GRU-based methods. Commonsense knowledge~\cite{34}, psychological knowledge~\cite{49}, and cognitive theory of emotion~\cite{50} are also used to enhance dialogue emotion recognition.

\textbf{Emotional Dialogue.} Emotional dialogue aims to generate emotional responses with two main strategies. One strategy is to specify a target emotion in advance~\cite{7,8,9,12}. The advantage of this advantage is that the generated emotions are flexible and controllable, and its disadvantage is that large-scale emotion-annotated dialogue corpora are required. The other strategy is to utilize the dialogue context to learn emotions by itself~\cite{6}, which is close to empathetic dialogue~\cite{11,13,46,53,54,55} supposing that listeners can infer speakers’ emotions~\cite{27}. The advantage of this strategy is that it can utilize the existing large-scale dialogue corpora, and its disadvantage is that the emotions of generated responses are challenging to control. Furthermore, a promising task emotional support dialogue~\cite{52} has recently emerged, which provides valuable assistance to people in need~\cite{51, 56, 57}.

\textbf{Controllable Text Generation.} Controllable text generation aims to generate texts with controllable styles. Style is defined as tokens belonging to a specific category or label~\cite{15}. Typical processes include training a large-scale conditional generation model from scratch, fine-tuning from a pre-trained language model, and replacing the key n-tuple to adjust the style of the generated sentence~\cite{23}. As a kind of style, emotion has good practical significance for its controllable generation. Emotion controlled text generation is to redefine the text to contain the specific emotion without changing the contextual intention~\cite{35,25,37,39,40}.

The differences between the proposed conversational agent and existing methods are: 

(1) As far as we know, we are the first to study a two-stage emotional response generation paradigm in the field of emotional dialogue specially. 

(2) We refine the response with dynamically recognized dialogue context emotions. However, current rewriting methods do not consider the dynamic acquisition of emotions. 

\section{Conclusions}
This paper designed a two-stage conversational agent in the field of emotional dialogue that generates content-related and emotional responses. The proposed conversational agent generates a semantically coherent prototype response in the first stage and emotionally refines the prototype response in the second stage. Extensive automatic and human evaluations have demonstrated that the proposed conversational agent can generate high-quality emotional responses of appropriate semantics, and reduce the demand for the sizeable emotion-annotated dialogue corpus.

In the future, to improve the proposed conversational
agent, we will explore the prediction of explicit and implicit expressions, and the flexible enhancement of other specific features besides the emotion of the dialogue system, such as domain and style adaptation of existing dialogue models.

\bibliographystyle{ACM-Reference-Format} 
\bibliography{sample}

\begin{thebibliography}{10}

\bibitem{6}
Nabiha Asghar, P.~Poupart, J.~Hoey, Xin Jiang, and Lili Mou.
\newblock Affective neural response generation.
\newblock {\em ArXiv}, abs/1709.03968, 2018.

\bibitem{14}
Laurie Carr, Marco Iacoboni, Marie-Charlotte Dubeau, John~C. Mazziotta, and
  Gian~Luigi Lenzi.
\newblock Neural mechanisms of empathy in humans: A relay from neural systems
  for imitation to limbic areas.
\newblock {\em Proceedings of the National Academy of Sciences},
  100(9):5497--5502, 2003.

\bibitem{28}
Chaotao Chen, Jinhua Peng, Fan Wang, Jun Xu, and Hua Wu.
\newblock Generating multiple diverse responses with multi-mapping and
  posterior mapping selection.
\newblock In {\em Proceedings of the Twenty-Eighth International Joint
  Conference on Artificial Intelligence, {IJCAI-19}}, pages 4918--4924.
  International Joint Conferences on Artificial Intelligence Organization, 7
  2019.

\bibitem{23}
Sumanth Dathathri, Andrea Madotto, Janice Lan, Jane Hung, Eric Frank, Piero
  Molino, Jason Yosinski, and Rosanne Liu.
\newblock Plug and play language models: A simple approach to controlled text
  generation, 2019.

\bibitem{17}
Jacob Devlin, Ming-Wei Chang, Kenton Lee, and Kristina Toutanova.
\newblock {BERT}: Pre-training of deep bidirectional transformers for language
  understanding.
\newblock In {\em Proceedings of the 2019 Conference of the North {A}merican
  Chapter of the Association for Computational Linguistics: Human Language
  Technologies, Volume 1 (Long and Short Papers)}, pages 4171--4186,
  Minneapolis, Minnesota, June 2019. Association for Computational Linguistics.

\bibitem{45}
Angela Fan, Mike Lewis, and Yann Dauphin.
\newblock Hierarchical neural story generation.
\newblock {\em arXiv preprint arXiv:1805.04833}, 2018.

\bibitem{frijda1987emotion}
Nico~H Frijda.
\newblock Emotion, cognitive structure, and action tendency.
\newblock {\em Cognition and emotion}, 1(2):115--143, 1987.

\bibitem{53}
Jun Gao, Yuhan Liu, Haolin Deng, Wei Wang, Yu~Cao, Jiachen Du, and Ruifeng Xu.
\newblock Improving empathetic response generation by recognizing emotion cause
  in conversations.
\newblock In Marie{-}Francine Moens, Xuanjing Huang, Lucia Specia, and
  Scott~Wen{-}tau Yih, editors, {\em Findings of the Association for
  Computational Linguistics: {EMNLP} 2021, Virtual Event / Punta Cana,
  Dominican Republic, 16-20 November, 2021}, pages 807--819. Association for
  Computational Linguistics, 2021.

\bibitem{34}
Deepanway Ghosal, Navonil Majumder, Alexander Gelbukh, Rada Mihalcea, and
  Soujanya Poria.
\newblock {COSMIC}: {CO}mmon{S}ense knowledge for e{M}otion identification in
  conversations.
\newblock In {\em Findings of the Association for Computational Linguistics:
  EMNLP 2020}, pages 2470--2481, Online, November 2020. Association for
  Computational Linguistics.

\bibitem{20}
Deepanway Ghosal, Navonil Majumder, Soujanya Poria, Niyati Chhaya, and
  Alexander Gelbukh.
\newblock {D}ialogue{GCN}: A graph convolutional neural network for emotion
  recognition in conversation.
\newblock In {\em Proceedings of the 2019 Conference on Empirical Methods in
  Natural Language Processing and the 9th International Joint Conference on
  Natural Language Processing (EMNLP-IJCNLP)}, pages 154--164, Hong Kong,
  China, November 2019. Association for Computational Linguistics.

\bibitem{5}
Sayan Ghosh, Mathieu Chollet, Eugene Laksana, Louis-Philippe Morency, and
  Stefan Scherer.
\newblock Affect-{LM}: A neural language model for customizable affective text
  generation.
\newblock In {\em Proceedings of the 55th Annual Meeting of the Association for
  Computational Linguistics (Volume 1: Long Papers)}, pages 634--642,
  Vancouver, Canada, July 2017. Association for Computational Linguistics.

\bibitem{33}
Devamanyu Hazarika, Soujanya Poria, Rada Mihalcea, Erik Cambria, and Roger
  Zimmermann.
\newblock {ICON}: Interactive conversational memory network for multimodal
  emotion detection.
\newblock In {\em Proceedings of the 2018 Conference on Empirical Methods in
  Natural Language Processing}, pages 2594--2604, Brussels, Belgium,
  October-November 2018. Association for Computational Linguistics.

\bibitem{32}
Devamanyu Hazarika, Soujanya Poria, Amir Zadeh, Erik Cambria, Louis-Philippe
  Morency, and Roger Zimmermann.
\newblock Conversational memory network for emotion recognition in dyadic
  dialogue videos.
\newblock In {\em Proceedings of the 2018 Conference of the North {A}merican
  Chapter of the Association for Computational Linguistics: Human Language
  Technologies, Volume 1 (Long Papers)}, pages 2122--2132, New Orleans,
  Louisiana, June 2018. Association for Computational Linguistics.

\bibitem{44}
Ari Holtzman, Jan Buys, Li~Du, Maxwell Forbes, and Yejin Choi.
\newblock The curious case of neural text degeneration.
\newblock {\em arXiv preprint arXiv:1904.09751}, 2019.

\bibitem{50}
Dou Hu, Lingwei Wei, and Xiaoyong Huai.
\newblock Dialoguecrn: Contextual reasoning networks for emotion recognition in
  conversations.
\newblock In Chengqing Zong, Fei Xia, Wenjie Li, and Roberto Navigli, editors,
  {\em Proceedings of the 59th Annual Meeting of the Association for
  Computational Linguistics and the 11th International Joint Conference on
  Natural Language Processing, {ACL/IJCNLP} 2021, (Volume 1: Long Papers),
  Virtual Event, August 1-6, 2021}, pages 7042--7052. Association for
  Computational Linguistics, 2021.

\bibitem{35}
Zhiting Hu, Zichao Yang, Xiaodan Liang, Ruslan Salakhutdinov, and Eric~P. Xing.
\newblock Toward controlled generation of text.
\newblock In Doina Precup and Yee~Whye Teh, editors, {\em Proceedings of the
  34th International Conference on Machine Learning}, volume~70 of {\em
  Proceedings of Machine Learning Research}, pages 1587--1596. PMLR, 06--11 Aug
  2017.

\bibitem{39}
N.~Keskar, Bryan McCann, L.~Varshney, Caiming Xiong, and R.~Socher.
\newblock Ctrl: A conditional transformer language model for controllable
  generation.
\newblock {\em ArXiv}, abs/1909.05858, 2019.

\bibitem{21}
Diederik~P. Kingma and Jimmy Ba.
\newblock Adam: A method for stochastic optimization, 2017.

\bibitem{49}
Jiangnan Li, Zheng Lin, Peng Fu, and Weiping Wang.
\newblock Past, present, and future: Conversational emotion recognition through
  structural modeling of psychological knowledge.
\newblock In Marie{-}Francine Moens, Xuanjing Huang, Lucia Specia, and
  Scott~Wen{-}tau Yih, editors, {\em Findings of the Association for
  Computational Linguistics: {EMNLP} 2021, Virtual Event / Punta Cana,
  Dominican Republic, 16-20 November, 2021}, pages 1204--1214. Association for
  Computational Linguistics, 2021.

\bibitem{28-diversity}
Jiwei Li, Michel Galley, Chris Brockett, Jianfeng Gao, and Bill Dolan.
\newblock A diversity-promoting objective function for neural conversation
  models.
\newblock In {\em Proceedings of the 2016 Conference of the North {A}merican
  Chapter of the Association for Computational Linguistics: Human Language
  Technologies}, pages 110--119, San Diego, California, June 2016. Association
  for Computational Linguistics.

\bibitem{46}
Qintong Li, Hongshen Chen, Zhaochun Ren, Pengjie Ren, Zhaopeng Tu, and Zhumin
  Chen.
\newblock Empdg: Multi-resolution interactive empathetic dialogue generation.
\newblock In Donia Scott, N{\'{u}}ria Bel, and Chengqing Zong, editors, {\em
  Proceedings of the 28th International Conference on Computational
  Linguistics, {COLING} 2020, Barcelona, Spain (Online), December 8-13, 2020},
  pages 4454--4466. International Committee on Computational Linguistics, 2020.

\bibitem{55}
Qintong Li, Piji Li, Zhaochun Ren, Pengjie Ren, and Zhumin Chen.
\newblock Knowledge bridging for empathetic dialogue generation.
\newblock 2022.

\bibitem{12}
Shifeng Li, Shi Feng, Daling Wang, Kaisong Song, Yifei Zhang, and Weichao Wang.
\newblock Emoelicitor: An open domain response generation model with user
  emotional reaction awareness.
\newblock In Christian Bessiere, editor, {\em Proceedings of the Twenty-Ninth
  International Joint Conference on Artificial Intelligence, {IJCAI-20}}, pages
  3637--3643. International Joint Conferences on Artificial Intelligence
  Organization, 7 2020.
\newblock Main track.

\bibitem{26}
Yanran Li, Hui Su, Xiaoyu Shen, Wenjie Li, Ziqiang Cao, and Shuzi Niu.
\newblock {D}aily{D}ialog: A manually labelled multi-turn dialogue dataset.
\newblock In {\em Proceedings of the Eighth International Joint Conference on
  Natural Language Processing (Volume 1: Long Papers)}, pages 986--995, Taipei,
  Taiwan, November 2017. Asian Federation of Natural Language Processing.

\bibitem{11}
Zhaojiang Lin, Andrea Madotto, Jamin Shin, Peng Xu, and Pascale Fung.
\newblock {M}o{EL}: Mixture of empathetic listeners.
\newblock In {\em Proceedings of the 2019 Conference on Empirical Methods in
  Natural Language Processing and the 9th International Joint Conference on
  Natural Language Processing (EMNLP-IJCNLP)}, pages 121--132, Hong Kong,
  China, November 2019. Association for Computational Linguistics.

\bibitem{29}
Chia-Wei Liu, Ryan Lowe, Iulian Serban, Mike Noseworthy, Laurent Charlin, and
  Joelle Pineau.
\newblock How {NOT} to evaluate your dialogue system: An empirical study of
  unsupervised evaluation metrics for dialogue response generation.
\newblock In {\em Proceedings of the 2016 Conference on Empirical Methods in
  Natural Language Processing}, pages 2122--2132, Austin, Texas, November 2016.
  Association for Computational Linguistics.

\bibitem{52}
Siyang Liu, Chujie Zheng, Orianna Demasi, Sahand Sabour, Yu~Li, Zhou Yu, Yong
  Jiang, and Minlie Huang.
\newblock Towards emotional support dialog systems.
\newblock In Chengqing Zong, Fei Xia, Wenjie Li, and Roberto Navigli, editors,
  {\em Proceedings of the 59th Annual Meeting of the Association for
  Computational Linguistics and the 11th International Joint Conference on
  Natural Language Processing, {ACL/IJCNLP} 2021, (Volume 1: Long Papers),
  Virtual Event, August 1-6, 2021}, pages 3469--3483. Association for
  Computational Linguistics, 2021.

\bibitem{13}
Navonil Majumder, Pengfei Hong, Shanshan Peng, Jiankun Lu, Deepanway Ghosal,
  Alexander Gelbukh, Rada Mihalcea, and Soujanya Poria.
\newblock {MIME}: {MIM}icking emotions for empathetic response generation.
\newblock In {\em Proceedings of the 2020 Conference on Empirical Methods in
  Natural Language Processing (EMNLP)}, pages 8968--8979, Online, November
  2020. Association for Computational Linguistics.

\bibitem{31}
Navonil Majumder, Soujanya Poria, Devamanyu Hazarika, Rada Mihalcea, Alexander
  Gelbukh, and Erik Cambria.
\newblock Dialoguernn: An attentive rnn for emotion detection in conversations.
\newblock {\em Proceedings of the AAAI Conference on Artificial Intelligence},
  33(01):6818--6825, Jul. 2019.

\bibitem{1}
Bilyana Martinovski and David~R. Traum.
\newblock Breakdown in human-machine interaction: the error is the clue.
\newblock In {\em ISCA tutorial and research workshop on Error handling in
  dialogue systems}, August 2003 2003.

\bibitem{24}
Courtney Napoles, Keisuke Sakaguchi, Matt Post, and Joel Tetreault.
\newblock Gleu without tuning, 2016.

\bibitem{40}
Kartikey Pant, Yash Verma, and R.~Mamidi.
\newblock Sentiinc: Incorporating sentiment information into sentiment transfer
  without parallel data.
\newblock {\em Advances in Information Retrieval}, 12036:312 -- 319, 2020.

\bibitem{27-bleu}
Kishore Papineni, Salim Roukos, Todd Ward, and Wei-Jing Zhu.
\newblock {B}leu: a method for automatic evaluation of machine translation.
\newblock In {\em Proceedings of the 40th Annual Meeting of the Association for
  Computational Linguistics}, pages 311--318, Philadelphia, Pennsylvania, USA,
  July 2002. Association for Computational Linguistics.

\bibitem{48}
Kishore Papineni, Salim Roukos, Todd Ward, and Wei{-}Jing Zhu.
\newblock Bleu: a method for automatic evaluation of machine translation.
\newblock In {\em Proceedings of the 40th Annual Meeting of the Association for
  Computational Linguistics, July 6-12, 2002, Philadelphia, PA, {USA}}, pages
  311--318. {ACL}, 2002.

\bibitem{2}
Timo Partala and V.~Surakka.
\newblock The effects of affective interventions in human-computer interaction.
\newblock {\em Interact. Comput.}, 16:295--309, 2004.

\bibitem{56}
Wei Peng, Yue Hu, Luxi Xing, Yuqiang Xie, Yajing Sun, and Yunpeng Li.
\newblock Control globally, understand locally: A global-to-local hierarchical
  graph network for emotional support conversation, 2022.

\bibitem{57}
Wei Peng, Ziyuan Qin, Yue Hu, Yuqiang Xie, and Yunpeng Li.
\newblock Fado: Feedback-aware double controlling network for emotional support
  conversation.
\newblock {\em Knowledge-Based Systems}, page 110340, 2023.

\bibitem{30}
Soujanya Poria, Erik Cambria, Devamanyu Hazarika, Navonil Majumder, Amir Zadeh,
  and Louis-Philippe Morency.
\newblock Context-dependent sentiment analysis in user-generated videos.
\newblock In {\em Proceedings of the 55th Annual Meeting of the Association for
  Computational Linguistics (Volume 1: Long Papers)}, pages 873--883,
  Vancouver, Canada, July 2017. Association for Computational Linguistics.

\bibitem{15}
Shrimai Prabhumoye, Alan~W Black, and Ruslan Salakhutdinov.
\newblock Exploring controllable text generation techniques.
\newblock In {\em Proceedings of the 28th International Conference on
  Computational Linguistics}, pages 1--14, Barcelona, Spain (Online), December
  2020. International Committee on Computational Linguistics.

\bibitem{3}
Helmut Prendinger, Junichiro Mori, and Mitsuru Ishizuka.
\newblock Using human physiology to evaluate subtle expressivity of a virtual
  quizmaster in a mathematical game.
\newblock {\em International Journal of Human-Computer Studies},
  62(2):231--245, 2005.
\newblock Subtle expressivity for characters and robots.

\bibitem{16}
Alec Radford, Jeff Wu, Rewon Child, David Luan, Dario Amodei, and Ilya
  Sutskever.
\newblock Language models are unsupervised multitask learners.
\newblock 2019.

\bibitem{27}
Hannah Rashkin, Eric~Michael Smith, Margaret Li, and Y-Lan Boureau.
\newblock Towards empathetic open-domain conversation models: A new benchmark
  and dataset.
\newblock In {\em Proceedings of the 57th Annual Meeting of the Association for
  Computational Linguistics}, pages 5370--5381, Florence, Italy, July 2019.
  Association for Computational Linguistics.

\bibitem{47}
Sahand Sabour, Chujie Zheng, and Minlie Huang.
\newblock {CEM:} commonsense-aware empathetic response generation.
\newblock {\em CoRR}, abs/2109.05739, 2021.

\bibitem{54}
Sahand Sabour, Chujie Zheng, and Minlie Huang.
\newblock {CEM:} commonsense-aware empathetic response generation.
\newblock {\em CoRR}, abs/2109.05739, 2021.

\bibitem{4}
M.~Skowron.
\newblock Affect listeners: Acquisition of affective states by means of
  conversational systems.
\newblock In {\em COST 2102 Training School}, 2009.

\bibitem{9}
Zhenqiao Song, Xiaoqing Zheng, Lu~Liu, Mu~Xu, and Xuanjing Huang.
\newblock Generating responses with a specific emotion in dialog.
\newblock In {\em Proceedings of the 57th Annual Meeting of the Association for
  Computational Linguistics}, pages 3685--3695, Florence, Italy, July 2019.
  Association for Computational Linguistics.

\bibitem{25}
A.~Sudhakar, Bhargav Upadhyay, and A.~Maheswaran.
\newblock Transforming delete, retrieve, generate approach for controlled text
  style transfer.
\newblock {\em ArXiv}, abs/1908.09368, 2019.

\bibitem{tian-etal-2020-skep}
Hao Tian, Can Gao, Xinyan Xiao, Hao Liu, Bolei He, Hua Wu, Haifeng Wang, and
  feng wu.
\newblock {SKEP}: Sentiment knowledge enhanced pre-training for sentiment
  analysis.
\newblock In {\em Proceedings of the 58th Annual Meeting of the Association for
  Computational Linguistics}, pages 4067--4076, Online, July 2020. Association
  for Computational Linguistics.

\bibitem{43-skep}
Hao Tian, Can Gao, Xinyan Xiao, Hao Liu, Bolei He, Hua Wu, Haifeng Wang, and
  Feng Wu.
\newblock {SKEP}: Sentiment knowledge enhanced pre-training for sentiment
  analysis.
\newblock In {\em Proceedings of the 58th Annual Meeting of the Association for
  Computational Linguistics}, pages 4067--4076, Online, July 2020. Association
  for Computational Linguistics.

\bibitem{51}
Quan Tu, Yanran Li, Jianwei Cui, Bin Wang, Ji{-}Rong Wen, and Rui Yan.
\newblock {MISC:} {A} mixed strategy-aware model integrating {COMET} for
  emotional support conversation.
\newblock {\em CoRR}, abs/2203.13560, 2022.

\bibitem{19}
Ashish Vaswani, Noam Shazeer, Niki Parmar, Jakob Uszkoreit, Llion Jones,
  Aidan~N Gomez, \L~ukasz Kaiser, and Illia Polosukhin.
\newblock Attention is all you need.
\newblock In I.~Guyon, U.~V. Luxburg, S.~Bengio, H.~Wallach, R.~Fergus,
  S.~Vishwanathan, and R.~Garnett, editors, {\em Advances in Neural Information
  Processing Systems}, volume~30. Curran Associates, Inc., 2017.

\bibitem{22}
Thomas Wolf, Lysandre Debut, Victor Sanh, Julien Chaumond, Clement Delangue,
  Anthony Moi, Pierric Cistac, Tim Rault, Remi Louf, Morgan Funtowicz, Joe
  Davison, Sam Shleifer, Patrick von Platen, Clara Ma, Yacine Jernite, Julien
  Plu, Canwen Xu, Teven Le~Scao, Sylvain Gugger, Mariama Drame, Quentin Lhoest,
  and Alexander Rush.
\newblock Transformers: State-of-the-art natural language processing.
\newblock In {\em Proceedings of the 2020 Conference on Empirical Methods in
  Natural Language Processing: System Demonstrations}, pages 38--45, Online,
  October 2020. Association for Computational Linguistics.

\bibitem{37}
Xing Wu, Tao Zhang, Liangjun Zang, Jizhong Han, and Songlin Hu.
\newblock Mask and infill: Applying masked language model for sentiment
  transfer.
\newblock In {\em IJCAI}, 2019.

\bibitem{36}
Jingjing Xu, X.~Sun, Qi~Zeng, Xuancheng Ren, X.~Zhang, Houfeng Wang, and W.~Li.
\newblock Unpaired sentiment-to-sentiment translation: A cycled reinforcement
  learning approach.
\newblock In {\em ACL}, 2018.

\bibitem{38}
Yi~Zhang, Jingjing Xu, Pengcheng Yang, and X.~Sun.
\newblock Learning sentiment memories for sentiment modification without
  parallel data.
\newblock In {\em EMNLP}, 2018.

\bibitem{26-generating}
Yizhe Zhang, Michel Galley, Jianfeng Gao, Zhe Gan, Xiujun Li, Chris Brockett,
  and Bill Dolan.
\newblock Generating informative and diverse conversational responses via
  adversarial information maximization.
\newblock In S.~Bengio, H.~Wallach, H.~Larochelle, K.~Grauman, N.~Cesa-Bianchi,
  and R.~Garnett, editors, {\em Advances in Neural Information Processing
  Systems}, volume~31. Curran Associates, Inc., 2018.

\bibitem{18}
Yizhe Zhang, Siqi Sun, Michel Galley, Yen-Chun Chen, Chris Brockett, Xiang Gao,
  Jianfeng Gao, Jingjing Liu, and Bill Dolan.
\newblock {DIALOGPT} : Large-scale generative pre-training for conversational
  response generation.
\newblock In {\em Proceedings of the 58th Annual Meeting of the Association for
  Computational Linguistics: System Demonstrations}, pages 270--278, Online,
  July 2020. Association for Computational Linguistics.

\bibitem{10}
Peixiang Zhong, D.~Wang, and C.~Miao.
\newblock An affect-rich neural conversational model with biased attention and
  weighted cross-entropy loss.
\newblock In {\em AAAI}, 2019.

\bibitem{7}
Hao Zhou, Minlie Huang, T.~Zhang, Xiaoyan Zhu, and Bing Liu.
\newblock Emotional chatting machine: Emotional conversation generation with
  internal and external memory.
\newblock In {\em AAAI}, 2018.

\bibitem{8}
Xianda Zhou and William~Yang Wang.
\newblock {M}oji{T}alk: Generating emotional responses at scale.
\newblock In {\em Proceedings of the 56th Annual Meeting of the Association for
  Computational Linguistics (Volume 1: Long Papers)}, pages 1128--1137,
  Melbourne, Australia, July 2018. Association for Computational Linguistics.

\end{thebibliography}


\end{document}